\documentclass[11pt]{article}

\usepackage[final]{acl}

\usepackage{times}
\usepackage{latexsym}
\usepackage[T1]{fontenc}
\usepackage[utf8]{inputenc}
\usepackage{microtype}
\usepackage{inconsolata}

\usepackage{graphicx}

\usepackage{booktabs}
\usepackage{multirow}

\usepackage{amsmath}
\usepackage{amsthm}
\usepackage{amssymb}
\usepackage{subcaption}
\usepackage{xcolor}
\usepackage[most]{tcolorbox}

\newtheorem{insight}{Insight}

\title{Distilling the Shape of Uncertainty: \\
What On-Policy Distillation Learns and How to Predict It}

\author{
 \textbf{Shiping Yang\textsuperscript{1 2}\thanks{Work done during an internship at Borealis AI}},
 \textbf{Chuanqi Tang\textsuperscript{1}},
 \textbf{Hossein Hajimirsadeghi\textsuperscript{2}},
 \textbf{Layla El Asri\textsuperscript{2}}, \\
 \textbf{Mohammad Amin Shabani\textsuperscript{2}},
 \textbf{Angel X. Chang\textsuperscript{1}},
\\
 \textsuperscript{1}Simon Fraser University  \ \
 \textsuperscript{2}Borealis AI  \ \
 \\
}

\begin{document}
\maketitle

\begin{abstract}

On-policy distillation (OPD) is increasingly adopted in the post-training of large language models. However, a stronger teacher may fail to improve the student. Prior work attributes this to a lack of new knowledge, but what that knowledge is remains unclear. We show that OPD learns not only \textit{how to branch}, i.e., which candidate tokens to consider, but also \textit{where to branch}, i.e., where to keep reasoning paths open. The existing token-overlap metric captures only the former. We introduce the entropy shape, the sequence of next-token entropies, to characterize the latter, and propose the Entropy Shape Correlation (ESC) to measure student-teacher agreement in entropy shape. Tracking ESC throughout training and intervening on a teacher's entropy while preserving its token preferences, we empirically show that \textit{where to branch} causally drives OPD gains, independently of \textit{how to branch}. Finally, we formulate OPD outcome prediction as a teacher selection task and establish an evaluation protocol for it, under which pre-OPD ESC selects the best teacher for both students and outperforms existing predictors. The code is available at the \href{https://github.com/maybenotime/opd-esc}{provided link}.

\end{abstract}

\section{Introduction}
\label{sec:intro}

\begin{figure*}[t]
  \centering
  \includegraphics[width=\textwidth]{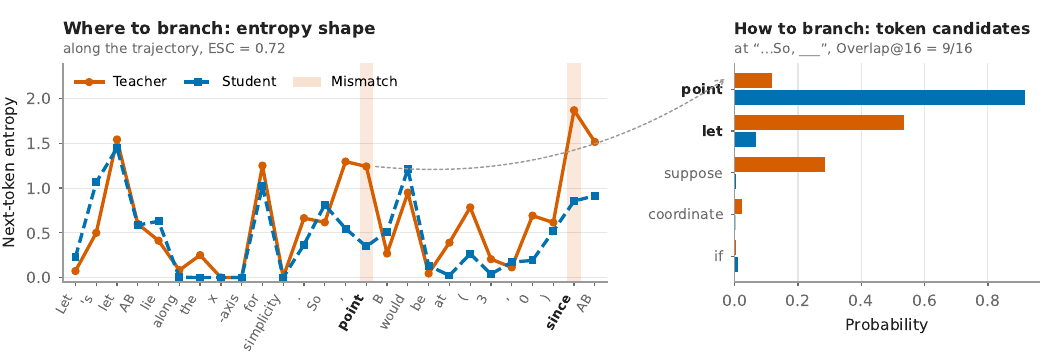}
  \caption{Illustration of where and how to branch on a student rollout (student: R1-Distill-1.5B; teacher: JustRL-1.5B). \textbf{Left}: \textit{where to branch}. Shaded positions mark mismatches, where the teacher tends to branch (high entropy) while the student does not (low entropy). \textbf{Right}: \textit{how to branch}. At the first mismatch, the student follows the current path with "point", whereas the teacher spreads probability over "let" and "suppose", typical fork tokens that open alternative reasoning paths. \textbf{Bold} tokens appear among the top-16 candidates of both models. ESC and Overlap@k measure student-teacher agreement on where and how to branch, respectively.}
  \label{fig:teaser}
\end{figure*}

On-policy distillation (OPD) has recently emerged as a standard post-training stage for large language models (LLMs)~\citep{xiao2026mimo,yang2026learning}. Unlike off-policy distillation, which trains the student on fixed teacher-generated sequences, OPD samples trajectories from the student itself and uses a teacher model to score every token~\citep{agarwal2024policy,gu2024minillm}. The student thus learns on the states it actually visits at inference time, avoiding the distribution mismatch of off-policy training~\citep{bengio2015scheduled,ranzato2015sequence}, while receiving dense token-level feedback instead of the sparse outcome reward of reinforcement learning (RL)~\citep{yang2026modularized}.

Despite these advantages, OPD still faces a long-standing issue of distillation: a stronger teacher does not guarantee better distillation~\citep{cho2019efficacy}. In off-policy distillation, this issue is typically attributed to a large distribution gap between the student and teacher~\citep{zhang-etal-2023-lifting,li2025small}. This explanation has motivated methods that select or construct teachers for compatibility with the student rather than for performance alone~\citep{xu-etal-2025-stronger,ding2025micota,zhang2025find}. For OPD, recent work shows that a post-trained teacher can yield better distillation than a stronger teacher trained with the same recipe as the student, and attributes this to new knowledge that the teacher acquires through the additional post-training~\citep{li2026rethinking}. However, it remains unclear what this new knowledge is, and such analyses only diagnose OPD post hoc. This leaves two open questions:\textit{what does the student learn from the teacher in OPD, and can we predict a teacher's gain before training?}

To understand what OPD learns, we first test whether the new knowledge is simply the teacher's ability to solve more training problems. We find that OPD gains are unaffected by whether the teacher solves these problems, suggesting that the new knowledge is a general reasoning behavior the teacher follows on both correct and incorrect trajectories. Since this behavior is acquired from post-training, we draw on recent findings that RL post-training shapes reasoning behavior mainly through a small set of high-entropy fork tokens, where the reasoning trajectory branches~\citep{wang2026beyond}. We therefore hypothesize that the new knowledge lies in how the teacher acts at these forks. Specifically, OPD teaches the student \textit{where to branch}, i.e., at which positions to keep multiple reasoning paths open rather than commit to one, and \textit{how to branch}, i.e., which candidate tokens to consider at each fork (Figure~\ref{fig:teaser}). Prior work tracks only the latter during OPD training, through the overlap of the student's and teacher's top-k tokens (Overlap@k)~\citep{li2026rethinking}. We characterize the former with the entropy shape (or uncertainty shape), the sequence of next-token entropies along a trajectory: high-entropy positions steer the reasoning path, while low-entropy positions follow it. We then measure student-teacher alignment in entropy shape with the Entropy Shape Correlation (ESC).

We provide both observational and causal evidence that \textit{where to branch} is a key component of what OPD learns. In the training dynamics, the student's entropy shape progressively aligns with the teacher's in runs that successfully improve the student, and ESC rises alongside the student's performance, whereas in failing runs ESC starts high and quickly plateaus. To test whether this alignment causally drives OPD gains, we construct counterfactual teachers by transplanting the entropy shape of one teacher onto another, which changes where a teacher branches while preserving which candidates it prefers. Experimental results show that a successful teacher's entropy shape alone improves the student, and that ESC can rise independently of Overlap@k, indicating that \textit{where to branch} is not a byproduct of \textit{how to branch}.

To predict a teacher's gain before training, we use ESC to measure how far the student's entropy shape is from the teacher's before OPD begins, as OPD gains come from closing this gap. We formulate this prediction as a teacher selection task, in which each predictor ranks the candidate teachers for a given student. We evaluate predictors by their rank correlation with the realized gains and by the regret of the selected teacher. In our experiments, pre-OPD ESC ranks the teachers correctly and selects the best one for both students, outperforming baseline predictors based on teacher performance, teacher size, and Overlap@k.

Our contributions can be summarized as follows: \textbf{1)} We identify \emph{where to branch} as a key component of what OPD learns, characterize it with the entropy shape, and propose ESC to measure student-teacher alignment in entropy shape. \textbf{2)} We empirically validate, by tracking ESC during training and constructing counterfactual teachers, that \emph{where to branch} drives OPD gains as a learning signal distinct from \emph{how to branch}. \textbf{3)} We formulate OPD outcome prediction as a teacher selection task and establish an initial evaluation suite for it, on which pre-OPD ESC outperforms existing predictors.
\section{Related work}
\label{sec:related}

\paragraph{On-policy distillation.}
Exposure bias in off-policy distillation motivated the recent interest in on-policy distillation. MiniLLM~\citep{gu2024minillm} introduces OPD for LLMs, optimizing the reverse KL objective on student-generated rollouts via policy gradients, while GKD~\citep{agarwal2024policy} generalizes this approach by directly minimizing token-level divergences on mixtures of on- and off-policy data. Thinking Machines Lab~\citep{lu2025onpolicydistillation} demonstrate that on-policy dense supervision is more efficient than outcome-reward RL and shows promise for continual learning. More recently, OPD has been extended to the self-distillation setting, where the student serves as its own teacher by conditioning on privileged information~\citep{hubotter2026reinforcement,shenfeld2026self,zhao2026self}.
Existing analyses explain what OPD learns through the progressive alignment of the student’s and teacher’s top-k candidates~\citep{li2026rethinking}, which reflects \textit{how to branch}. In this work, we further identify \textit{where to branch} as a distinct component of the OPD learning signal.

\paragraph{Curse of capacity gap.}
A large teacher–student capacity gaps can diminish or even reverse the benefits of distillation~\citep{cho2019efficacy,mirzadeh2020improved}. This curse of capacity gap has been documented in language model distillation~\citep{zhang-etal-2023-lifting}, instruction tuning~\citep{xu-etal-2025-stronger}, and chain-of-thought distillation~\citep{li2025small}. One line of work addresses it through better teacher selection~\citep{zhang2025towards,busbridge2025distillation}. Another line of research bridges the teacher-student gap by constructing more suitable teachers. For example,~\citet{ding2025micota} employ intermediate-sized models as teacher assistants, and~\citet{zhang2025find} achieve personalized distillation through a query-level teacher router.
These studies, however, focus on off-policy distillation. OPD exhibits a similar phenomenon, where a stronger teacher may fail to improve the student, yet its cause remains underexplored.
\section{Preliminaries}
\label{sec:prelim}

\subsection{Notation}
\label{sec:prelim:notation}
Let $x \in \mathcal{D}_x$ denote a query and $y = (y_1, \ldots, y_T)$ be a response of length $T$ . We write $y_{<t} = (y_1, \ldots, y_{t-1})$ for the prefix up to step $t$. A language model $\pi$ defines a next-token distribution $\pi(\cdot \mid x, y_{<t})$ over a vocabulary $\mathcal{V}$. We refer to the model being trained as the \emph{student} $\pi_\theta$, with learnable parameters $\theta$, and to
a frozen reference model as the \emph{teacher} $\pi_T$. Both models share the same vocabulary $\mathcal{V}$. Knowledge distillation transfers knowledge from $\pi_T$ to $\pi_\theta$ by minimizing the KL divergence between the distributions $\mathcal{P}$ and $\mathcal{Q}$:
\begin{equation}
  D_{\mathrm{KL}}(\mathcal{P} \,\|\, \mathcal{Q})
  = \sum_{v \in \mathcal{V}} \mathcal{P}(v) \log \frac{\mathcal{P}(v)}{\mathcal{Q}(v)}.
  \label{eq:kl}
\end{equation}

\subsection{On-policy distillation}
\label{sec:prelim:opd}

\paragraph{Objective.}
Given $x \sim \mathcal{D}_x$, OPD samples a response $y \sim \pi_\theta(\cdot \mid x)$ and forwards it through both models, yielding $p_t = \pi_\theta(\cdot \mid x,y_{<t})$ and $q_t = \pi_T(\cdot \mid x,y_{<t})$ at every prefix. The standard OPD objective is the sequence-level reverse KL divergence between the student and teacher under trajectories sampled from the student policy, which, by autoregressive factorization, can be formalized as a token-level KL objective:

{\small
\begin{equation}
\mathcal{L}_{\mathrm{OPD}}(\theta)
= \mathbb{E}_{x \sim \mathcal{D}_x,\; y \sim \pi_\theta(\cdot \mid x)}
\left[\sum_{t=1}^{T} D_{\mathrm{KL}}\!\left(p_t \,\|\, q_t\right)\right].
\label{eq:opd-token}
\end{equation}}

In practice, OPD implementations differ in how they compute or estimate each token-level KL term: (1) \emph{full-vocabulary OPD} computes the exact KL divergence between the student and teacher distributions over the entire vocabulary; (2) \emph{top-$k$ OPD} reduces the cost by restricting both distributions to the $k$ tokens with the highest student probabilities and renormalizing them; and (3) \emph{sampled-token OPD} evaluates only the token sampled by the student, yielding an unbiased Monte Carlo estimator of each token-level KL term.

\paragraph{Sampled-token OPD.}
By evaluating only the token sampled by the student, sampled-token OPD is the most lightweight variant and has therefore become the most commonly used implementation in prior on-policy distillation work~\citep{lu2025onpolicydistillation,xiao2026mimo,yang2026learning}. \textbf{Unless otherwise specified, all our experiments use this variant.}
Given $y_t \sim p_t$, we define the per-token log-ratio:
\begin{equation}
\Delta_t \triangleq \log q_t(y_t) - \log p_t(y_t).
\label{eq:per-token-ratio}
\end{equation}
Since
$\mathbb{E}_{y_t \sim p_t}[\Delta_t]
= -D_{\mathrm{KL}}(p_t \,\|\, q_t)$,
each $\Delta_t$ is an unbiased single-sample estimator of the negative token-level reverse KL, so maximizing the expected $\Delta_t$ is equivalent to minimizing Eq.~\eqref{eq:opd-token}.
The sign of $\Delta_t$ sets the direction of the update: a positive value reinforces the sampled token, which the teacher assigns higher probability than the student does, while a negative value suppresses a token the student overweights relative to the teacher. The magnitude of $\Delta_t$ reflects how far the two models disagree on that token.

\paragraph{Policy optimization.}
Following \citet{lu2025onpolicydistillation}, OPD is implemented via policy gradient with $\Delta_t$ as the per-token advantage. To stabilize updates, we apply clipping~\citep{schulman2017proximal} to the importance ratio and maximize the clipped surrogate objective:

{\small
\begin{equation}
\mathcal{J}_{\mathrm{clip}}(\theta)
= \mathbb{E}_t\!\left[
\min\!\left(
r_t \bar{a}_t,\;
\mathrm{clip}(r_t, 1-\epsilon, 1+\epsilon)\,\bar{a}_t
\right)\right],
\label{eq:opd-clip}
\end{equation}
}
\noindent where $r_t = \pi_\theta(y_t \mid x, y_{<t}) /
\pi_{\theta_{\mathrm{old}}}(y_t \mid x, y_{<t})$
is the importance ratio and $\bar{a}_t$ is the advantage derived from $\Delta_t$ with gradients stopped. 

\subsection{Dynamic metrics}
\label{sec:dynamic_metrics}

The following metrics are monitored throughout OPD training in subsequent experiments and are both computed on the student’s rollouts. Let $p_t$ and $q_t$ denote the student's and teacher's next-token distributions at step $t$, and let $H^{p}_t$ and $H^{q}_t$ be their entropies.

\paragraph{Overlap@k.}
Introduced by \citet{li2026rethinking}, this metric quantifies the alignment
between the student’s and teacher’s candidate spaces. We define their top-$k$
token sets at position $t$ as
$S_t^{(p)} = \operatorname{TopK}(p_t,k)$ and
$S_t^{(q)} = \operatorname{TopK}(q_t,k)$, respectively. It is computed as the
average proportion of tokens shared by the two sets:
\begin{equation}
  \mathrm{Overlap@}k
  =
  \frac{1}{T}
  \sum_{t=1}^{T}
  \frac{\left|S_t^{(p)} \cap S_t^{(q)}\right|}{k}.
  \label{eq:overlap}
\end{equation}

A value close to 1 indicates that the student has located the teacher's high-probability region. \citet{li2026rethinking} identify the progressive alignment of this metric as a signature of successful OPD. We set $k=16$ following their setup.

\paragraph{Entropy gap.}
This metric measures the absolute difference between the student's and
teacher's mean per-position entropies:
\begin{equation}
  \Delta H
  =
  \left|
    \frac{1}{T}\sum_{t=1}^{T} H_t^p
    -
    \frac{1}{T}\sum_{t=1}^{T} H_t^q
  \right|.
  \label{eq:entropy-gap}
\end{equation}
By averaging over positions, $\Delta H$ captures only the difference in overall entropy level, discarding all positional information.

\section{What does on-policy distillation learn?}
\label{sec:learn}

Previous work shows that a post-trained teacher can yield better distillation than a stronger one following the same training recipe as the student~\citep{li2026rethinking}.
It attributes this to \textit{new knowledge} acquired through the additional RL training, but what this knowledge is remains underexplored. In this section, we further investigate the underlying mechanisms of on-policy distillation to answer this question. We first demonstrate that OPD does not depend on whether the teacher can solve the training problems (\S~\ref{sec:learn:data}), then propose the entropy shape as what OPD transfers (\S~\ref{sec:learn:shape}), and finally verify that this shape causally drives OPD's gains by constructing counterfactual teachers (\S~\ref{sec:learn:counterfactual_teacher}).

\subsection{OPD does not rely on correct supervision}
\label{sec:learn:data}

The post-trained teachers tested in \citet{li2026rethinking} are obtained by applying additional RL to the same-recipe teachers, and can therefore typically solve more of the training prompts.
This raise a natural hypothesis: On-policy distillation learn from correct supervision and the gap between two teachers in OPD is driven by their knowledge gap on the training data. We assess this hypothesis by examining whether the teacher's correctness on the training problems affects the outcome of OPD.

\paragraph{Setup.}
We study two students, DeepSeek-R1-Distill-Qwen-1.5B~\citep{guo2025deepseek} and Qwen3-1.7B (non-thinking)~\citep{yang2025qwen3}, each paired with a same-recipe teacher and a post-trained teacher. For the DeepSeek family, we compare DeepSeek-R1-Distill-Qwen-7B (R1-Distill-7B) and JustRL-DeepSeek-1.5B (JustRL-1.5B)~\citep{he2025justrl}, where the latter is obtained by applying RL to DeepSeek-R1-Distill-Qwen-1.5B on the DAPO-Math-17k dataset~\citep{yu2026dapo}. 
For the Qwen family, we compare Qwen3-8B (non-thinking) and Qwen3-4B-RL-Math~\citep{yang2026learning}, where the latter is obtained by applying RL to Qwen3-4B (non-thinking).
Using DeepMath-103K~\citep{he2026deepmath}, we construct two correctness-controlled subsets of 7,000 prompts each, where a prompt is labeled as solved by a teacher if a single sampled rollout reaches the correct final answer. The two subsets are: (1) \textbf{both-correct}, containing prompts that both teachers solve, and (2) \textbf{both-wrong}, containing prompts on which both teachers fail.
By construction, the effect of knowledge gap between different teachers is ablated. We then conduct OPD for every student-teacher pair on both subsets.
Unless otherwise specified, all experiments use the default setting described in Appendix ~\ref{app:training_details} and are evaluated on AIME 2024, AIME 2025, AMC2023~\citep{li2024numinamath}, HMMT February 2025, HMMT February 2026~\citep{dekoninck2026matharena} and BeyondAIME~\citep{bytedance_seed_2025_beyondaime}. We report average accuracy over 4 samples (avg@4) as the primary evaluation metric.

\paragraph{Results.}
As shown in Table~\ref{tab:correctness}, the post-trained teachers consistently outperform the same-recipe teachers in both families. Distillation from the post-trained teacher remains effective even when it fails to solve any training problem, while distillation from the same-recipe teacher remains ineffective even when it solves every training problem.
Therefore, the gap between teachers cannot be attributed to a knowledge gap on the training data.
Moreover, for a fixed teacher, the both-correct and both-wrong subsets yield comparable distillation outcomes across all student-teacher pairs, and the both-wrong subset even produces slightly larger gains.
These results suggest that OPD transfers a more general reasoning ability, and its success is independent of the correctness of the supervision.

\begin{table*}[t]
\centering
\small
\setlength{\tabcolsep}{4pt}
\resizebox{\textwidth}{!}{%
\begin{tabular}{llccccccc}
\toprule
Teacher & Training set & AIME24 & AIME25 & AMC & BeyondAIME & HMMT25 & HMMT26 & Average \\
\midrule
\multicolumn{9}{l}{\textit{Student: DeepSeek-R1-Distill-Qwen-1.5B \quad (average before OPD: 25.5)}} \\
\midrule
JustRL-1.5B       & both-wrong   & 39.1 & 23.3 & 74.7 & 13.2 & 20.8 & 18.2 & 31.6 {\scriptsize($+6.1$)} \\
JustRL-1.5B       & both-correct & 33.3 & 28.3 & 70.8 & 12.8 & 15.0 & 18.9 & 29.9 {\scriptsize($+4.4$)} \\
\addlinespace[2pt]
R1-Distill-7B     & both-wrong   & 27.5 & 25.0 & 60.2 &  8.8 & 15.8 & 15.9 & 25.5 {\scriptsize($+0.0$)} \\
R1-Distill-7B     & both-correct & 25.8 & 20.0 & 62.0 &  8.8 & 17.5 & 15.2 & 24.9 {\scriptsize($-0.6$)} \\
\midrule
\multicolumn{9}{l}{\textit{Student: Qwen3-1.7B (non-thinking) \quad (average before OPD: 13.0)}} \\
\midrule
Qwen3-4B-RL-Math  & both-wrong   & 29.1 & 22.5 & 62.7 & 11.2 & 15.8 & 18.9 & 26.7 {\scriptsize($+13.7$)} \\
Qwen3-4B-RL-Math  & both-correct & 29.1 & 22.5 & 63.3 & 11.8 & 15.0 & 16.7 & 26.4 {\scriptsize($+13.4$)} \\
\addlinespace[2pt]
Qwen3-8B          & both-wrong   & 14.2 &  7.5 & 41.9 &  4.5 &  8.3 &  9.1 & 14.2 {\scriptsize($+1.2$)} \\
Qwen3-8B          & both-correct & 15.0 & 10.0 & 38.6 &  4.0 &  5.0 &  6.8 & 13.2 {\scriptsize($+0.2$)} \\
\bottomrule
\end{tabular}}
\caption{OPD results on correctness-controlled subsets. Each student is distilled from two teachers on the both-wrong and both-correct subsets (7k prompts each). Average is the mean avg@4 over the six benchmarks, followed by its absolute change relative to the student's performance before OPD.}
\label{tab:correctness}
\end{table*}


\subsection{OPD learns where and how to branch}
\label{sec:learn:shape}

We argue that what OPD learns is a general reasoning behavior intrinsic to the teacher that goes beyond surface-level knowledge. This behavior is transferred to the student through OPD regardless of whether the teacher’s reasoning trajectory ultimately reaches the correct answer. 
Inspired by the recent finding that reasoning gains from RL are driven by a small set of high-entropy fork tokens at reasoning branch points~\citep{wang2026beyond}, we hypothesize that OPD operates similarly: by supervising student rollouts with the RL post-trained teacher’s token-level distributions, OPD teaches the student where to branch during reasoning and what alternatives to consider at each branch point.

\paragraph{Where to branch: entropy shape.}

\citet{li2026rethinking} tracks OPD’s alignment progress using the $\mathrm{overlap@}k$ metric in \S~\ref{sec:dynamic_metrics}, which measures the overlap between the teacher’s and student’s top-$k$ highest-probability tokens. This token-level metric primarily captures \emph{how to branch}: which candidates are considered at each step. However, because it treats each position independently, it does not characterize \emph{where to branch} along the reasoning trajectory. 
To capture this trajectory-level structure, we introduce the \emph{entropy shape} (or \emph{uncertainty shape}), defined as the sequence of next-token entropies along a reasoning trajectory. High-entropy regions mark points where the model considers multiple plausible continuations, while lower-entropy regions indicate commitment to a particular reasoning path. We visualize the entropy shapes of the student and teacher before and after OPD in Figure~\ref{fig:entropy-shape}, which shows that their entropy shapes become aligned after distillation.

\paragraph{Metrics.}
Let $H_t^p = H(p_t)$ and $H_t^q = H(q_t)$ denote the entropies of the
student's and teacher's next-token distributions, respectively, at position
$t \in \{1,\ldots,T\}$. We compare their uncertainty shapes along the same trajectory using \textbf{Entropy Shape Correlation (ESC)}, defined as the Pearson correlation
between the two entropy sequences:
\begin{equation}
  \rho
  =
  \mathrm{Pearson}\!\left(
    \{H_t^p\}_{t=1}^{T},
    \{H_t^q\}_{t=1}^{T}
  \right).
  \label{eq:rho}
\end{equation}
ESC measures whether the student’s and teacher’s entropy profiles rise and fall in sync, thereby capturing their agreement on where to branch along the reasoning trajectory. Since the Pearson correlation is invariant to shifting and positive scaling of either sequence, ESC excludes the effect of differences in absolute entropy levels and measures the alignment of entropy shape alone.
For comparison, we also track two metrics defined in \S~\ref{sec:dynamic_metrics}: $\mathrm{Overlap@16}$, which measures agreement on candidate tokens, and the entropy gap $\Delta H$, which measures the difference in entropy value.

\paragraph{Setup.}
We use the same student–teacher pairs as in \S~\ref{sec:learn:data}: DeepSeek-R1-Distill-Qwen-1.5B (R1-Distill-1.5B) distilled from JustRL-1.5B and R1-Distill-7B, and Qwen3-1.7B distilled from Qwen3-4B-RL-Math and Qwen3-8B. We run OPD on DAPO-Math-17K~\citep{yu2026dapo} and monitor ESC, Overlap@16, and entropy gap $\Delta H$ throughout training, together with the student's average performance across six benchmarks.

\begin{figure*}[t]
  \centering
  \includegraphics[width=\textwidth]{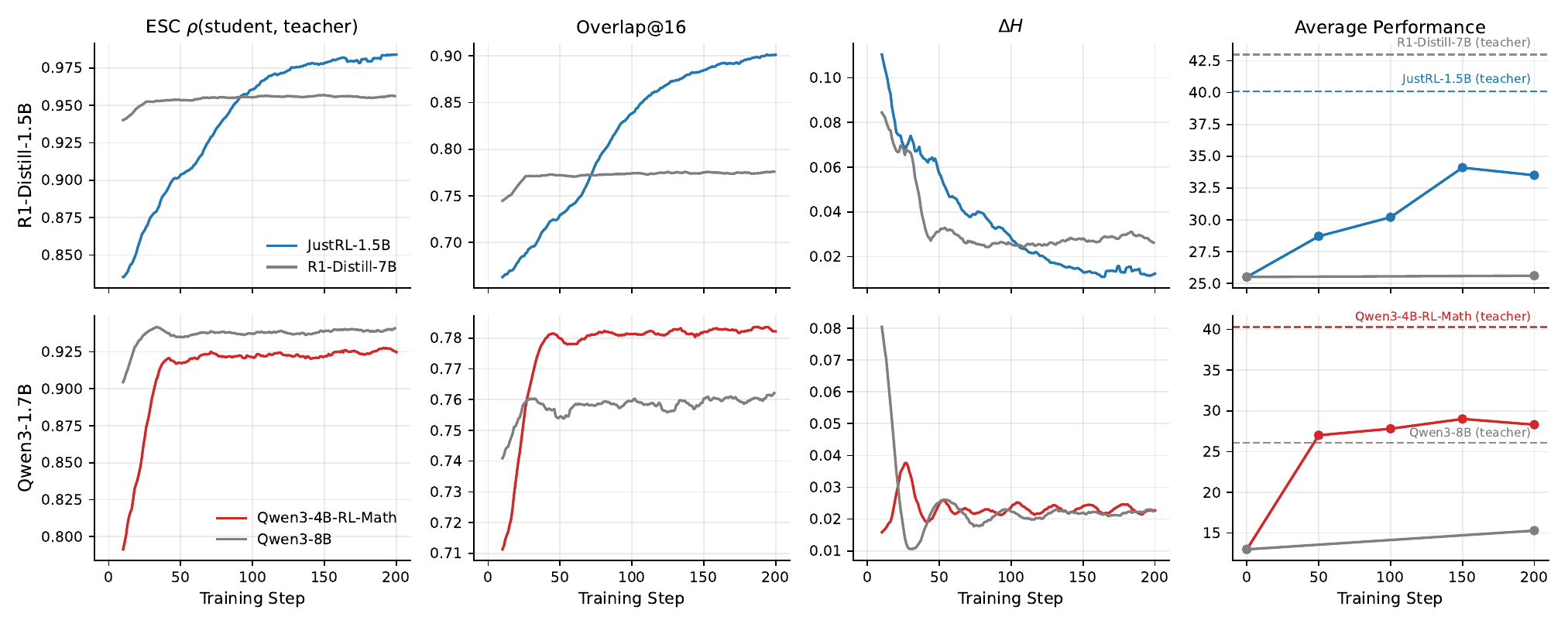}
  \caption{Training dynamics of successful and failing OPD runs. \textbf{Top:} R1-Distill-1.5B student. Successful runs are shown in color and failing runs in grey. The first three columns show ESC, Overlap@16, and $\Delta H$ over training (10-step moving average). The last column shows the average performance on the six benchmarks, with successful runs evaluated every 50 steps and failing runs evaluated only at the beginning and end of training; dashed lines indicate teacher performance. \textbf{Bottom:} the same layout for the Qwen3-1.7B student.
  }
  \label{fig:dynamics}
\end{figure*}

\paragraph{Results.}
Figure~\ref{fig:dynamics} shows sharply different dynamics between successful and failing OPD runs.
ESC rises substantially in successful OPD runs, while in failing runs it starts high and quickly plateaus, leaving little room for further alignment between the student’s and teacher’s entropy shapes.
Moreover, ESC closely tracks the student’s downstream performance: average performance rises alongside ESC and levels off once ESC saturates.
A similar pattern emerges for $\mathrm{Overlap@16}$, which increases in successful runs while remaining flat in failing ones.
These trends suggest that learning how to branch and learning where to branch largely progress together during OPD. 
By contrast, $\Delta H$ does not distinguish successful from failing runs, as all runs converge to similarly low values. Notably, in the successful Qwen3-1.7B run, $\Delta H$ first increases and eventually converges slightly above its initial value, even as the student’s performance improves substantially.
Additional experiments in \S~\ref{sec:signal} demonstrate that $\Delta H$ is uncorrelated with the performance gains from OPD. Together, the results
suggest that what matters for OPD is where the student is uncertain along the
trajectory, rather than how uncertain it is overall.
These observations are consistent with our hypothesis that OPD teaches the student where to branch. However, they remain correlational: the alignment of entropy shape could merely accompany OPD's gains without driving them. We further provide causal evidence in \S~\ref{sec:learn:counterfactual_teacher}.

\subsection{Causal analysis with counterfactual teachers}
\label{sec:learn:counterfactual_teacher}
So far, our experiments has shown only observational correlations.
The progressive entropy shape alignment we see with ESC may be a byproduct of successful OPD training rather than the \emph{cause} of its gains. 
To isolate the causal role of the entropy shape in OPD, we manipulate the teacher's entropy shape to study the impact of distillation with a different shape.
We construct counterfactual teachers by transplanting one teacher’s entropy shape onto another.

\paragraph{Transplanting entropy shape.}
To manipulate a teacher's entropy precisely, we use per-position temperature scaling. Let $z_{A,t}$ be the logits of teacher $A$ at position $t$ of a student rollout, and $H_{B,t}$ the entropy of teacher $B$ at the same position. We denote transplanting $B$'s entropy shape onto $A$ by $A \leftarrow B$. At each position, we solve
\begin{equation}
  H\!\left(\mathrm{softmax}(z_{A,t}/T_t)\right) = H_{B,t}
  \label{eq:transplant}
\end{equation}
for $T_t$ using Newton's method in $\log T$ with bracketing, and rescale $A$'s logits accordingly to obtain the counterfactual teacher logits $\widetilde{q}^{A \leftarrow B}_t = \mathrm{softmax}(z_{A,t}/T_t)$.
Since dividing logits by a positive scalar preserves their ranking, $\widetilde{q}^{A \leftarrow B}_t$ keeps $A$'s token preferences, including its argmax and top-$k$ sets, while matching $B$'s entropy at every position. During OPD, $\widetilde{q}^{A \leftarrow B}_t$ replaces the original teacher distribution, and all other settings remain unchanged.

\paragraph{Counterfactual teacher setup.}

Let $A$ be a successful teacher and $B$ a failing one.\footnote{We refer to a teacher as successful if OPD with it improves the student, and as failing otherwise.} We construct two types of counterfactual teachers: (1) $\widetilde{q}^{B \leftarrow A}$, which transplants the entropy shape of the successful teacher onto the failing one, to examine whether the entropy shape of a successful teacher alone is enough for OPD to yield gains; and (2) $\widetilde{q}^{A \leftarrow B}$, which transplants the entropy shape of the failing teacher onto the successful one, to examine how much the successful teacher is affected once it loses its original entropy shape.
Specifically, for the Qwen3-1.7B student, we set Qwen3-4B-RL-Math as the successful teacher $A$ and Qwen3-4B as the failing teacher $B$. For the R1-Distill-1.5B student, we set JustRL-1.5B as $A$ and the student itself as $B$, since distilling a model into itself yields zero divergence and thus no learning signal.
All runs are trained on DAPO-Math-17K and use the same training and evaluation settings as before.

\begin{figure}[t]
  \centering
  \includegraphics[width=\columnwidth]{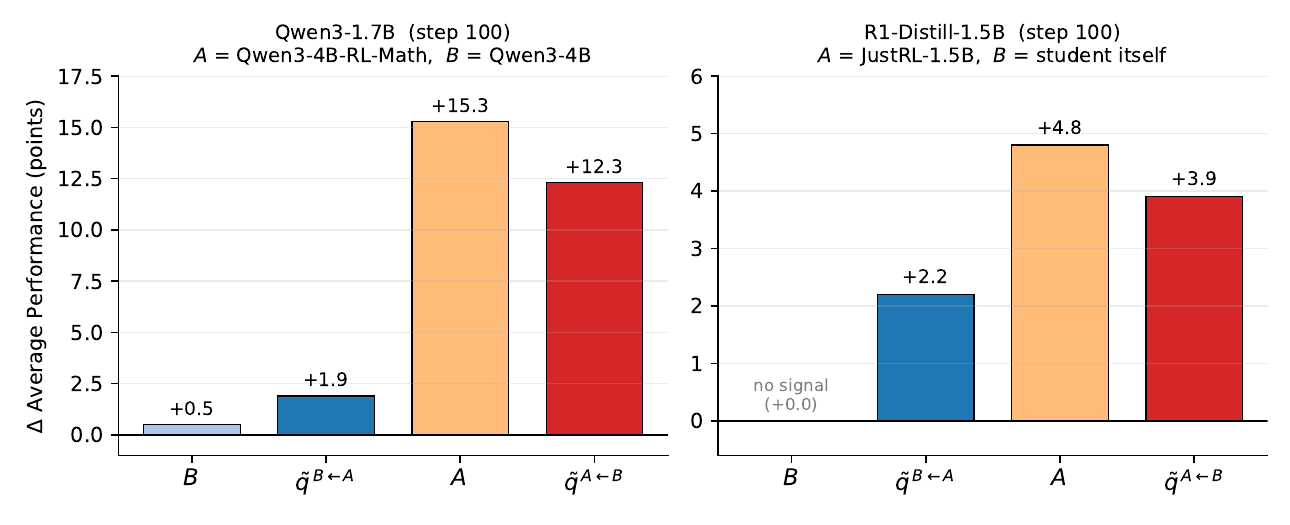}
  \caption{OPD gains with original and counterfactual teachers. 
  We study the effect of transplanting the entropy shape between a successful teacher $A$ and a failing teacher $B$ for two student models: Qwen3-1.7B (left) and R1-Distill-1.5B (right).
  We report the student's average gain over the base model on the six benchmarks after 100 OPD steps for the two teacher models ($A$, $B$) and the transplanted models. 
  $\widetilde{q}^{B \leftarrow A}$ transplants $A$'s entropy shape onto $B$; $\widetilde{q}^{A \leftarrow B}$ reverses the direction. 
  Transplanting successful teacher's entropy shape ($B \leftarrow A$) increases the performance gain, while transplanting failing teacher's entropy shape ($A \leftarrow B$) lowers the performance gain.
  For R1-Distill-1.5B, $B$ is the student itself and thus supplies no learning signal on its own.}
  \label{fig:counterfactual}
\end{figure}
\paragraph{Results.}
Figure~\ref{fig:counterfactual} provides causal evidence that entropy shape affects the outcome of OPD. Transplanting the successful teacher’s entropy shape onto the failing teacher improves performance in both model families. The effect is most evident for R1-Distill-1.5B: after 100 OPD steps, $\widetilde{q}^{B \leftarrow A}$ improves the student by $+2.2$ points, compared with $+4.8$ points from JustRL-1.5B. This gain is notable because the counterfactual teacher keeps only the student's own token preferences, which provide no learning signal without the transplant.
Conversely, replacing the successful teacher’s original entropy shape reduces the OPD gain in both model families. Yet most of the gain remains because the teacher’s token preferences are preserved, indicating that how to branch also provides a substantial learning signal. These results show that where to branch and how to branch jointly constitute the effective learning signal transferred through OPD.

\paragraph{Training dynamics of counterfactual teachers.}
As shown in Figure~\ref{fig:ct_dynamics}, each counterfactual run moves only one of the two metrics over the training: for $\widetilde{q}^{B \leftarrow A}$,
which carries the successful teacher's entropy shape, ESC against the shape source rises while Overlap@16 against the token preference source stays flat; for $\widetilde{q}^{A \leftarrow B}$, which carries the failing teacher's shape, the reverse holds. Where and how to branch are therefore separable, and
transferring either one alone can drive the gain.

\section{How to predict OPD outcomes}
\label{sec:signal}



\begin{table*}[t]
\centering
\small
\begin{tabular*}{\textwidth}{@{\extracolsep{\fill}}lcccccc@{}}
\toprule
 & \multicolumn{2}{c}{All ($n{=}10$)} & \multicolumn{2}{c}{$r_s$ per student} & \multicolumn{2}{c}{Regret per student} \\
\cmidrule(lr){2-3}\cmidrule(lr){4-5}\cmidrule(lr){6-7}
Predictor & $r_s$ & 95\% CI & R1-Distill-1.5B & Qwen3-1.7B & R1-Distill-1.5B & Qwen3-1.7B \\
\midrule
ESC (ours)          & $+0.98$ & $[+0.78,\,+1.00]$ & $+1.00$ & $+1.00$ & $0.0$ & $0.0$ \\
Overlap@16          & $+0.49$ & $[-0.31,\,+0.95]$ & $+0.30$ & $+1.00$ & $10.3$ & $0.0$ \\
$\Delta H$          & $+0.20$ & $[-0.54,\,+0.89]$ & $+0.70$ & $-0.10$ & $0.0$  & $6.5$ \\
Teacher performance & $-0.01$ & $[-0.56,\,+0.81]$ & $-0.60$ & $+1.00$ & $7.5$  & $0.0$ \\
Teacher size        & $-0.61$ & $[-0.76,\,+0.25]$ & $-0.60$ & $-0.70$ & $10.8$ & $10.2$ \\
\bottomrule
\end{tabular*}
\caption{Performance of OPD predictors. The \emph{All} columns pool all ten student--teacher pairings, whereas the \emph{$r_s$ per student} and \emph{Regret per student} columns compute the corresponding metrics separately within each student's five-teacher pool, better reflecting the practical setting of selecting a teacher for a given student.}
\label{tab:prediction}
\end{table*}

Beyond tracking the learning progress during training, ESC can also be computed before OPD starts (pre-OPD ESC), to predict which student--teacher pairs will have relatively high performance gain from OPD and to select an appropriate teacher for a given student.





\paragraph{Task formulation.}
Given a student and a pool of candidate teachers, the goal is to predict the potential downstream gain that OPD will produce before training, and to select the teacher that maximizes it. Formally, let $S$ be a student and $\mathcal{T} = \{T_1, \dots, T_N\}$ a pool of candidate teachers, and let $\Delta(S, T_i)$ denote the downstream gain of $S$ after OPD with $T_i$. A predictor $m$ assigns each pair a score $m(S, T_i)$, computed on a small probe set of prompts before training, and selects the teacher with the highest score:
\begin{equation}
  \hat{T} = \arg\max_{T_i \in \mathcal{T}} \, m(S, T_i).
  \label{eq:selection}
\end{equation}

\paragraph{Evaluation suite.}
We establish an initial evaluation suite for the proposed task. Due to the high cost of obtaining the ground-truth gain $\Delta(S,T_i)$, we only construct a small evaluation set of 10 student-teacher pairings. Specifically, we use R1-Distill-1.5B and Qwen3-1.7B as student and pair each with five diverse teachers for OPD training, using the resulting performance gain as the ground-truth $\Delta(S, T_i)$. Full details of the 10 data points are provided in Appendix~\ref{sec:app:teachers}.
To evaluate predictor performance, we consider the following metrics: 
(1) \textbf{Spearman correlation} $r_s$ between the predictor's ranking of the candidate teachers and their ranking by realized gain $\Delta$. A larger $|r_s|$ indicates a stronger association between the predictor score and the realized gain, regardless of direction. (2) \textbf{Regret}, defined as $\max_{T} \Delta(T) - \Delta(\hat T)$, where $\hat T$ is the teacher ranked first by the predictor. It measures the performance lost by selecting a teacher with the predictor instead of choosing the best teacher in the candidate pool. A regret of zero therefore indicates that the predictor selects the best teacher.(3) \textbf{95\% bootstrap confidence interval}. Given the small number of student--teacher pairings, we additionally report a confidence interval for the signed $r_s$ to assess the stability of the observed correlation. We obtain the interval by resampling the pairings with replacement $5{,}000$ times and taking the $2.5$th and $97.5$th percentiles of the recomputed correlations. An interval containing zero indicates that the observed association is not stable in direction under resampling.

\paragraph{Baselines}
We compare ESC against two groups of baselines, all of which can be applied before training. (1) \textbf{Student-teacher baselines}, which characterize the relationship between the student and teacher.
For this group, we use $\mathrm{Overlap@k}$ and the \emph{entropy gap} $\Delta H$ introduced in \S~\ref{sec:dynamic_metrics}. Lower $\mathrm{Overlap@k}$ and a larger $\Delta H$ are taken to predict larger gains. Both are computed on the same 1,000 probe rollouts as ESC. (2) \textbf{Teacher-only baselines}, which depend solely on properties of the teacher and require no student--teacher probe.  
We use the teacher's \emph{performance} on the evaluation benchmarks and its \emph{size} (parameter count) to predict gains, with higher values taken to predict larger gains for both cases.

\paragraph{Results.}
Figure~\ref{fig:entropy-correlation} visualizes the relationship between pre-OPD ESC and the realized
OPD gain $\Delta(S,T_i)$. We observe a strong negative association:
student-teacher pairs with lower pre-OPD ESC tend to yield larger gains.
Lower ESC indicates a larger entropy shape gap, leaving more headroom for OPD to align. Therefore, lower ESC values correspond to larger predicted gains.
Table~\ref{tab:prediction} compares ESC with four baselines on ten student--teacher pairings. ESC achieves the strongest overall correlation with realized OPD gains and is the only predictor with perfect rank correlation for both students ($r_s=1.00$). It also selects the best teacher for both students, yielding zero regret. In contrast, the other baselines are inconsistent across students, and all of their bootstrap intervals include zero.

\section{Conclusion}
\label{sec:conclusion}

In this work, we investigate what OPD transfers from a teacher to student. We identify two aspects of what OPD learns, \textit{where to branch} and \textit{how to branch}, and show that \textit{where to branch} causally drives OPD gains, independently of \textit{how to branch}. We measure student-teacher agreement on \textit{where to branch} with ESC, which enables teacher selection before training. Beyond this, ESC could serve as a monitor for early stopping, and since OPD aligns a student’s entropy shape with its teacher’s, it may also help detect whether one model has been distilled from another through OPD.

\section*{Limitations}

Due to limited computational resources, we study only the sampled-token variant of OPD, and all experiments are on math reasoning tasks, with 1.5B and 1.7B students and teachers of at most 8B parameters. Whether our findings hold for other OPD variants, in domains such as code generation or instruction following, or at larger scales therefore remains open. In addition, our evaluation suite for teacher selection contains only ten student-teacher pairings, five for each of two students, since each data point requires a full OPD run and evaluation on six benchmarks. With so few pairings, the reported correlations have wide confidence intervals, so the results should be viewed as preliminary evidence rather than a rigorous evaluation of predictors.




\bibliography{custom}

\clearpage
\appendix
\section{Training and Evaluation Details}
\label{app:training_details}

\begin{table}[h]
\centering
\small
\begin{tabular}{lc}
\toprule
\textbf{Item} & \textbf{Value} \\
\midrule
Training temperature   & 1.0 \\
Top-$p$                & 1.0 \\
Global batch size      & 64 \\
Mini batch size        & 64 \\
Rollout number         & 1 \\
Teacher LogProb top-$K$ & 64 \\
Max prompt length      & 1024 \\
Max response length    & 16384 \\
Optimizer              & AdamW \\
Learning rate          & 1e-6 \\
Clip ratio $\epsilon$  & 0.2 \\
KL coefficient         & 0.0 \\
Training steps         & 200 \\
Save frequency         & 50 \\
\bottomrule
\end{tabular}
\caption{Default hyperparameters for OPD.}
\label{tab:hparams}
\end{table}

\paragraph{Evaluation.}
All models are evaluated with vLLM at temperature $0.7$, top-$p$ $0.8$, top-$k$ $20$, with four samples per problem. The maximum generation length is $16{,}384$ tokens for R1-Distill-1.5B (thinking mode) and $8{,}192$ tokens for Qwen3-1.7B (non-thinking mode). We report the unweighted mean over the six competitions.

\section{Teachers in the Evaluation Suite}
\label{sec:app:teachers}

Table~\ref{tab:teachers} lists the ten student--teacher pairings of the
prediction task with the teachers' Hugging Face identifiers and the realized
six-benchmark gain at step $200$.

\begin{table}[h]
\centering
\small
\setlength{\tabcolsep}{4pt}
\resizebox{\columnwidth}{!}{%
\begin{tabular}{llr}
\toprule
Teacher & Hugging Face id & $\Delta$ \\
\midrule
\multicolumn{3}{l}{\textit{Student: DeepSeek-R1-Distill-Qwen-1.5B}} \\
JustRL-1.5B    & \texttt{hbx/JustRL-DeepSeek-1.5B}                     & $+8.1$  \\
Nemotron-1.5B  & \texttt{nvidia/Nemotron-Research-Reasoning-Qwen-1.5B} & $+11.0$ \\
R1-Distill-7B  & \texttt{deepseek-ai/DeepSeek-R1-Distill-Qwen-7B}      & $+0.2$  \\
Skywork-OR1-7B & \texttt{Skywork/Skywork-OR1-Math-7B}                  & $+3.5$  \\
Light-R1-7B    & \texttt{qihoo360/Light-R1-7B-DS}                      & $+0.7$  \\
\midrule
\multicolumn{3}{l}{\textit{Student: Qwen3-1.7B}} \\
Qwen3-4B-RL-Math   & \texttt{Keven16/Qwen3-4B-Non-Thinking-RL-Math-Step500}  & $+15.3$ \\
Qwen3-4B-Code-1200 & \texttt{Keven16/Qwen3-4B-Non-Thinking-RL-Code-Step1200} & $+8.8$  \\
Qwen3-8B-GRPO-MATH & \texttt{xiaoyuchen1/Qwen3-8B-GRPO-MATH}                 & $+5.1$  \\
Qwen3-8B           & \texttt{Qwen/Qwen3-8B}                                  & $+2.3$  \\
Qwen3-4B           & \texttt{Qwen/Qwen3-4B}                                  & $+0.5$  \\
\bottomrule
\end{tabular}}
\caption{Teachers used in the prediction task and the six-benchmark average gain ($\Delta$) of the student.}
\label{tab:teachers}
\end{table}

\section{Entropy Shape Before and After OPD}
\label{sec:app:shapefig}

\begin{figure*}[t]
  \centering
  \includegraphics[width=\textwidth]{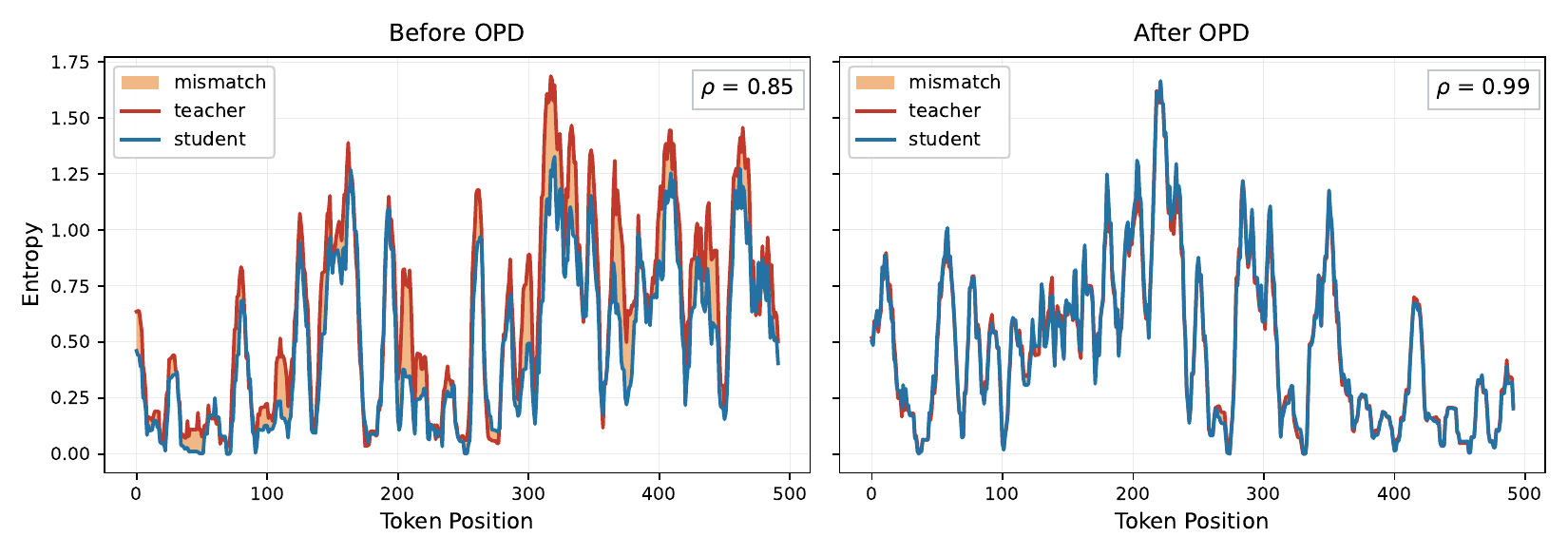}
  \caption{Per-position entropy ($9$-token moving average) of DeepSeek-R1-Distill-1.5B and its JustRL-1.5B teacher on the student's own rollout, before (left, $\rho=0.85$) and after OPD (right,
    $\rho=0.99$); shading marks the mismatch.}
  \label{fig:entropy-shape}
\end{figure*}

Figure~\ref{fig:entropy-shape} shows the student's and the teacher's entropy
shapes on a student rollout, before and after OPD.

\section{Pre-OPD ESC Against the Realized Gain}
\label{sec:app:scatter}

\begin{figure}[h]
  \centering
  \includegraphics[width=\columnwidth]{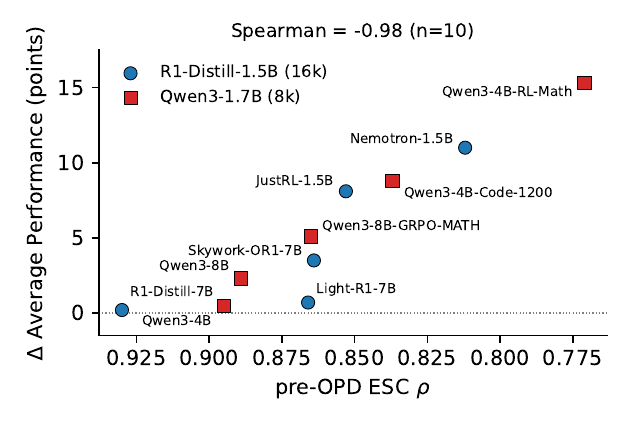}
  \caption{Pre-OPD ESC against the six-benchmark average gain for the ten student-teacher pairings in \S\ref{sec:signal}. Pairings with lower pre-OPD ESC leave more room for entropy shape alignment and yield larger gains.}
  \label{fig:entropy-correlation}
\end{figure}

Figure~\ref{fig:entropy-correlation} plots the pre-OPD ESC of each student-teacher pairing against its realized OPD gain.

\section{Training Dynamics of Counterfactual Teachers}
\label{sec:app:ctdyn}

\begin{figure*}[t]
  \centering
  \includegraphics[width=\textwidth]{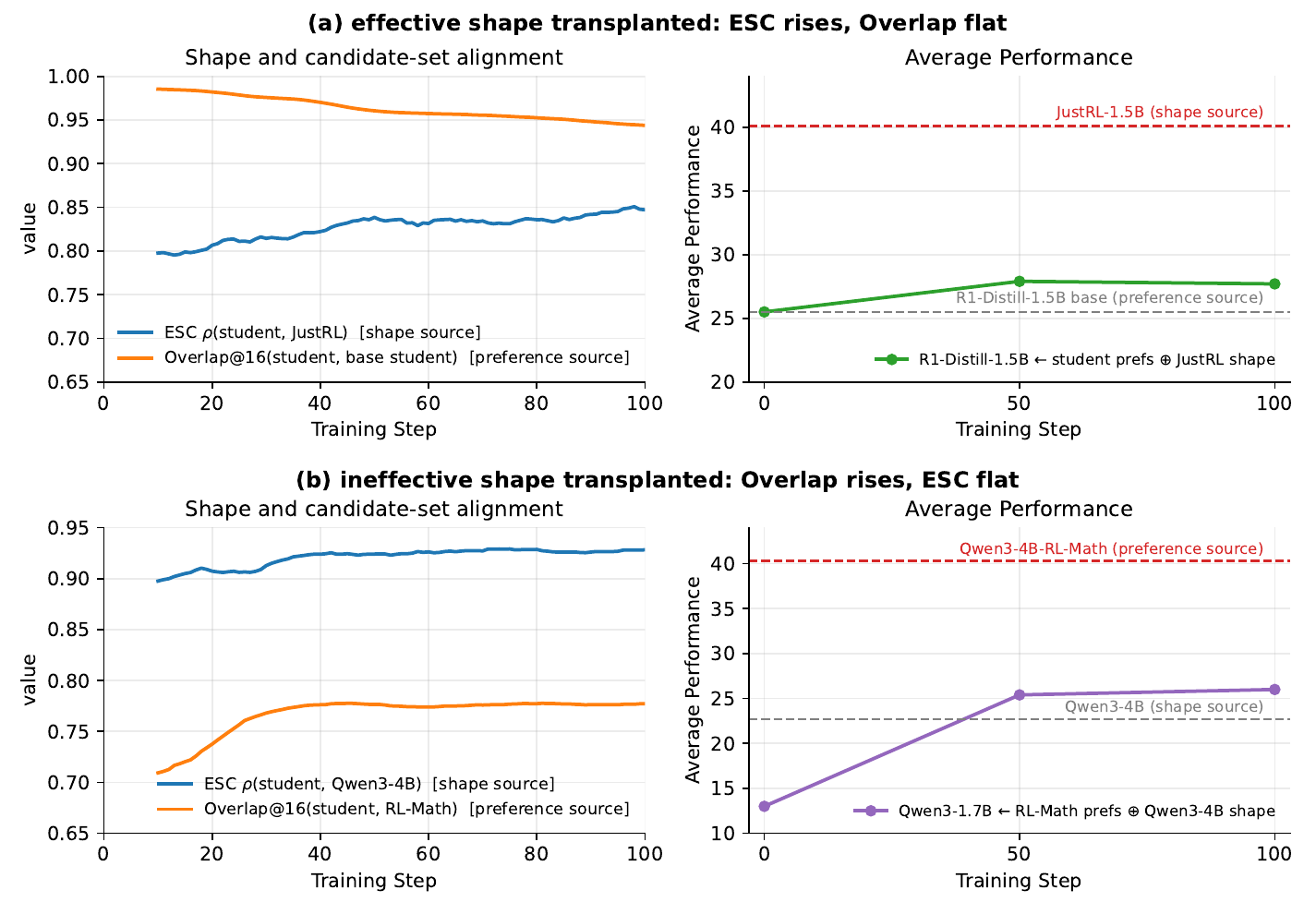}
  \caption{\textbf{Training dynamics of counterfactual teachers.} Each row tracks one of the counterfactual runs of Figure~\ref{fig:counterfactual}. \emph{Left:} ESC between the student and
    the model that supplies the entropy shape, and Overlap@16 between the
    student and the model that provides the token preferences. \emph{Right:} six-benchmark average accuracy at saved checkpoints, with dashed lines marking the two source models. \textbf{(a)} R1-Distill-1.5B
    distilled from $\widetilde{q}^{B \leftarrow A}$, with preferences from the
    student itself and shape from JustRL-1.5B. \textbf{(b)} Qwen3-1.7B distilled from
    $\widetilde{q}^{A \leftarrow B}$, with preferences from Qwen3-4B-RL-Math
    and shape from Qwen3-4B. Where and how to branch can thus be learned either jointly or independently.}
  \label{fig:ct_dynamics}
\end{figure*}

Figure~\ref{fig:ct_dynamics} tracks the two counterfactual runs of
\S\ref{sec:learn:counterfactual_teacher} over their first 100 training steps.

\end{document}